\documentclass[12pt]{article}
\usepackage[dvips, usenames]{color}
\usepackage{titletoc}
\usepackage{latexsym}
\usepackage{amsmath}
\usepackage{amssymb}
\usepackage{multicol}

\usepackage{enumerate}
\usepackage{natbib}

\usepackage{graphicx}
\usepackage{subfigure}
\usepackage{indentfirst}
\usepackage{epsfig}
\usepackage{mathrsfs}
\usepackage{epsfig,graphics,subfigure,psfrag,amsmath,amssymb}
\usepackage{indentfirst,latexsym,bm}
\usepackage{float}
\usepackage{setspace}
\usepackage{booktabs}
\usepackage{indentfirst}
\usepackage{amssymb}
\usepackage{amsthm}
\usepackage{graphicx}
\usepackage{subfigure}
\usepackage{bm}
\newtheorem{theorem}{Theorem}

\newtheorem{definition}{Definition}
\newtheorem{assumption}{Assumption}
\newtheorem{lemma}{Lemma}
\newtheorem{remark}{Remark}
\usepackage{epstopdf}
\allowdisplaybreaks

\newcommand{\blind}{0}

\begin{document}

\def\spacingset#1{\renewcommand{\baselinestretch}%
{#1}\small\normalsize} \spacingset{1}


\if0\blind
{
  \title{\bf Maximum Correntropy Criterion Regression models with tending-to-zero scale parameters}
  \author{Ying Jing,\quad Lianqiang Yang\\
    \ \\
    \textit{School of Mathematical Sciences, Anhui University, Hefei, China}\\
    }
  \maketitle
} \fi

\if1\blind
{
  \bigskip
  \bigskip
  \bigskip
  \begin{center}
    {\LARGE\bf Title}
\end{center}
  \medskip
} \fi

\bigskip
\begin{abstract}
Maximum correntropy criterion regression (MCCR) models have been well studied within the frame of statistical learning when the scale parameters take fixed values or go to infinity. This paper studies the MCCR models with tending-to-zero scale parameters. It is revealed that the optimal learning rate of MCCR models is ${\mathcal{O}}(n^{-1})$ in the asymptotic sense when the sample size $n$ goes to infinity. In the case of finite samples, the performances on robustness of MCCR, Huber and the least square regression models are compared. The applications of these three methods on real data are also displayed.
\end{abstract}

\noindent%
\spacingset{1.45}
\section{Introduction}
\label{intro}
It is known that the classical least square regression models achieve the optimal efficiency when the noises are Gaussian, however, they always underperform if the data is contaminated by non-Gaussian noises or outliers. Some robust regression models have been well developed in the past decades such as the median regression, the modal regression, the Huber regression and the least trimmed squares regression, etc.
Moreover, a new robust regression model named the maximum correntropy criterion regression (MCCR) has been theoretically studied within the frame of statistical learning in \cite{2015Learning}. Correntropy is constructed based on a kernel function and it is a generalized similarity measure between two random variables (see \cite{2006Generalized,gunduz2009correntropy,2007Correntropy,He2011Robust,2012Maximum} and \cite{2016Correntropy}). Correntropy induced loss function $\sigma^{2}(1-e^{-\frac{t^2}{\sigma^2}})$ and its scale parameter $\sigma$ are used to control the trade-off between the convergence rate and the robustness of the estimated regression function in MCCR models. While the scale parameter goes to infinity, the convergence rate of estimator has been established under some moment conditions or some absolute bounded conditions on the response variable (see Theorem 4 and 5 in \cite{2015Learning}), while the scale parameter is a fixed constant large enough, the convergence rate has been presented under the symmetric and uniformly bounded noises (see Theorem 6 in \cite{2015Learning}). However, the case that the scale parameter goes to zero has not been considered.

Some other works of MCCR have been presented. The model with the mixed symmetric stable noises and a fixed scale parameter was discussed in \cite{2018Learning}. The situation that the response variable with conditions of \begin{math} (1+\varepsilon)\end{math}-moment and the scale parameter greater than 1 was studied in \cite{feng2020learning}. Particularly, \cite{feng2020statistical} declared that the model with the correntropy induced loss function $\sigma^{-1}(1-e^{-\frac{t^2}{\sigma^2}})$ and a tending-to-zero scale parameter was modal regression. It should be noted that the loss functions for MCCR and the modal regression are essentially different when the scale parameters go to zero.

MCCR models with small or tending-to-zero scale parameters are worth to be considered. First, the scale parameters of MCCR play the same roles as the bandwidths of kernel density estimations, when the sample size $n$ is large or where the sample points are dense, the small or tending-to-zero scale parameters are expected to decrease the fitting error. Second, the scale parameters are similar to the smoothing parameters of local polynomial estimations, the small ones are needed where the regression functions have fluctuated. Last, \cite{2015Learning} has noted that the smaller scale parameters provide MCCR models with better robustness, it is meaningful to study the consistency of the estimates with nice robustness.

In this paper, we first give the theoretical study on MCCR models with tending-to-0 scale parameters. The convergence rates of estimates are discussed under the assumption of symmetric noises when the sample size goes to infinity. Then, the simulations and applications are presented to compare MCCR, Huber, and least square (LS) regression models as the sample size is finite. The rest of this paper is organized as follows. In Section 2, we give a short introduction to MCCR. The convergence rates of MCCR estimates with tending to 0 scale parameters under mixed symmetric noise are presented in Section 3. Section 4 and 5 display the numerical experiments and applications. The last is Section 6 for some conclusions.

\section{ Definition of MCCR}
We first assume that the regression model is given as
\begin{equation}
Y=f^{*}(X)+\varepsilon,
\end{equation}
where $X$ is the independent variable takes values in a compact metric space $\mathcal{X}$, the dependent variable $Y\in\mathcal{Y}=\mathbb{R}$, and $\varepsilon$ is the noise subject to $\mathbb{E}(\varepsilon|X)=0$. The purpose of a regression model is to learn the unknown regression function $f^*$ based on the given observations $z=\left\{(x_i,y_i)\right\}_{i=1}^n$, which are  independently sampled from the distribution $\rho$ of $(X, Y)$ on  $\mathcal{X} \times \mathcal{Y}$.

Given two scalar random variables $U$ and $V$, the correntropy $\mathcal{V}_\sigma$ between $U$ and $V$ is defined as $\mathcal{V}_{\sigma}(U,V)=\mathbb{E}{\mathcal{K}_{\sigma}(U,V)}$, where $\mathcal{K}_{\sigma}(\cdot,\cdot)$ is the Gaussian kernel function with the scale parameter $\sigma>0$. Theoretically, the idea of MCCR is to infer $f^*$  based on the maximization of $\mathcal{V}_\sigma$. Empirically, MCCR obtains the estimates of $f^*$ by maximizing the mean version of $\mathcal{V}_{\sigma}$ as follows

$$\hat{\mathcal{V}}_{\sigma,z}(f)=\frac{1}{n}{\sum_{i=1}^{n}{\mathcal{K}}_\sigma(y_i,f(x_i))}.$$
Let the output function of MCCR as
$$f_z=\arg\mathop{\max}\limits_{f\in \mathcal{H}}\hat{\mathcal{V}}_{\sigma,z}(f),$$
where the hypothesis space $\mathcal{H}$ is assumed to be a compact subset of a continuous functions space.
Furthermore, let the correntropy induced loss function $l_\sigma:\mathbb{R} \to [0,+\infty)$  as (see Definition 1 in \cite{2015Learning})
\begin{equation}
l_\sigma(t)=\sigma^{2}(1-e^{-\frac{t^2}{\sigma^2}}),
\end{equation}
then the estimator $f_z$ of MCCR can be equivalently expressed as
\begin{equation}
f_z=\arg\mathop{\min}\limits_{f\in \mathcal{H}}\frac{1}{n}{\sum_{i=1}^{n}l_\sigma(y_i-f(x_i))}.
\end{equation}
The discussions on the loss functions in \cite{2015Learning} and \cite{2013Robust} show that the scale parameter $\sigma$ tunes the tradeoff between robustness and convergence rate of MCCR model, a decreasing $\sigma$ enhances the robustness and reduces the convergence rate and vice versa.

\section{MCCR with mixture of symmetric noises}
In this section, we give the theoretical research on MCCR with a mixture of symmetric noises.
\subsection{Definitions and assumptions}
First, let $\rho_\mathcal{X}$, $\rho_\mathcal{Y}$ be the marginal distributions of $X, Y$ respectively. For any $f \in \mathcal{H}$, the empirical error in (3) is defined as
$${\varepsilon}_{z}^{\sigma}(f)=\frac{1}{n}{\sum_{i=1}^{n}l_\sigma(y_i-f(x_i))},$$
and its population version ${\varepsilon}^{\sigma}(f)$ is defined as
$${\varepsilon}^{\sigma}(f)=\int\limits_{\mathcal{X} \times \mathcal{Y}}l_{\sigma}(y-f(x))d{\rho}.$$
Let the distance between $f$ and $f^*$ under $L_{\rho_\mathcal{X}}^2$ be $\Vert f-f^*\Vert_{L_{\rho_\mathcal{X}}^2}^2$. Moreover, let notation $a_1 \lesssim a_2$ for $a_1, a_2\in\mathbb{R} $ means that there exists a positive constant $a_3$ such that $a_1 \leq a_{3}a_2$.
\begin{definition}[Mixture of symmetric distributions]
The univariate distribution function $P(t)$ (density function $p(t)$) is a mixed symmetric distribution if it is a convex combination of symmetric distributions $P_{i}(t)$ (density function $p_{i}(t)$), where $i=1, 2, \cdots, K$ and $K$ is a positive integer, that is, there exists $w_1, w_2,\cdots, w_K$ with $w_{i}>0, i=1, 2, \cdots, K$ and $\sum_{i=1}^{K}w_i=1$, such that for any $t\in \mathbb{R}$, there are
$$P(t)={\sum_{i=1}^{K}w_{i}P_{i}(t)},\quad p(t)={\sum_{i=1}^{K}w_{i}p_{i}(t)}.$$
\end{definition}
\begin{definition}[$l^{2}$-empirical covering number]
Let $x=\left\{x_1,x_2,\cdots,x_n\right\}\subset {\mathcal{X}}^n$ and  $\eta>0$. The $l^{2}$-empirical covering number $\mathcal{N}_{2}(\mathcal{H},\eta)$ of the hypothesis space $\mathcal{H}$ with radius $\eta$ is defined as follows (See \cite{anthony2009neural, guo2013concentration} and \cite{2007Learnability})
\[
\begin{aligned}
\mathcal{N}_{2}(\mathcal{H},\eta)&:=\sup\limits_{n\in \mathbb{N}} \sup\limits_{x\in {\chi}^n} \inf \left\{ l\in \mathbb{N}:\exists \left\{f_i\right\}_{i=1}^l \subset \mathcal{H}\, \mbox{such that for each}\, f\in \mathcal{H}\,\mbox{there exists some}\right.\\
&\left. i\in \left\{i=1,2,\cdots,l\right\}\, \mbox{with}\; \frac{1}{n}\sum_{j=1}^{n}\vert{f(x_{j})-f_{i}(x_j)\vert}^2\leq {\eta}^2 \right\}.
\end{aligned}
\]
\end{definition}
\begin{assumption}[Mixture of symmetric noises]
The noise $\varepsilon$ is a mixture of symmetric noises with 0 as the center, that is, for any $x\in \mathcal{X}$, the density $p_{\varepsilon|x}$ of the noise  $\varepsilon$ holds the following forms
$$p_{\varepsilon|x}(t)={\sum_{i=1}^{K}w_{i}p_{\varepsilon,i|x}(t)},\quad t\in \mathbb{R},$$
where $K$ is a positive integer, $w_{i}>0,i=1,2,\cdots,K$ with $\sum_{i=1}^{K}w_i=1$, and $p_{\varepsilon,i|x}$  is the density function of a symmetric distribution $P_{\varepsilon,i|x}$  with 0 as the center.
\end{assumption}

\begin{assumption}
 For any $x\in \mathcal{X}$ and $i=1,2,\cdots,K$, the Fourier transform $\widehat{p_{\varepsilon,i|x}}$  of $p_{\varepsilon,i|x}$  is positive, and there exist two positive constants $c_{0,i},C_{0,i}$, such that $\widehat{p_{\varepsilon,i|x}}(\xi)\geq C_{0,i}$ for $\xi \in [-c_{0,i},c_{0,i}]$.
\end{assumption}

\begin{assumption}[Complexity Assumption]
There exist positive constants $s$ and $c$ with $0<s<2$, such that
$$\log \mathcal{N}_2(\mathcal{H},\eta)\leq {c\eta}^{-s}, \forall \eta >0.$$
\end{assumption}

Moreover, we assume that $\sup\limits_{f\in \mathcal{H}}\Vert f\Vert_{\infty}\leq M$ and $\Vert f^*\Vert_{\infty}\leq M$, where $M$ is a positive constant. Assumptions 1 and 2 allow the noises to be kinds of random variables, such as Gaussian distribution, Cauchy distribution, Laplace distribution, Linnik distribution, symmetric stable distribution and the convex combinations of the above distributions (See \cite{1998Tails, fama1968some} and \cite{miller1978properties}). Assumption 3 is a common representation  for the complexity of  hypothesis space in statistical learning (See \cite{zhou2002covering} and \cite{cucker2007learning}).

\subsection{ Theoretical results on convergence rates}
We give two theorems on the learning efficiency of MCCR in this subsection.

\begin{theorem}
Suppose that Assumptions 1, 2 and $f^{*}\in \mathcal{H}$ hold. For a fixed scale parameter $\sigma>0$, we have
$$f^*=\mathop{\arg\min}_{f\in \mathcal{H}}{\varepsilon}^{\sigma}(f),$$
and for any $f\in \mathcal{H}$,
$$c_{\sigma}\Vert f-f^*\Vert_{L_{\rho_\mathcal{X}}^2}^2\leq {\varepsilon}^{\sigma}(f)-{\varepsilon}^{\sigma}(f^*)\leq \Vert f-f^*\Vert_{L_{\rho_\mathcal{X}}^2}^2,$$
where $c_{\sigma}$ is a positive constant which will be given explicitly in the  proof.
\end{theorem}

Theorem 1 shows that the population version estimator of MCCR can accurately represent the unknown function $f^*$ when $f^{*}\in \mathcal{H}$ and the noise is a mixture of symmetric distributions with positive Fourier transform and the Fourier transform has a lower bound of positive constant in a finite region. In this sense, $f_z$ can be regarded as an unbiased estimator of $f^*$. In addition, if let $f=f_z$, Theorem 1 shows that the excess risk of MCCR can be bounded by $L_{\rho_\mathcal{X}}^2$ distance between the MCCR estimator $f_z$ and the real unknown function $f^*$, which helps us to get Theorem 2.

\begin{theorem}
Suppose that Assumptions 1, 2, 3 and $f^{*}\in \mathcal{H}$ hold. Let $f_z$ be given in equation (3) and scale parameter $\sigma=n^{\theta}$. For any $0<{\delta}<1$, with probability at least $1-\delta$, it holds that
$$\Vert f_z-f^*\Vert_{L_{\rho_\mathcal{X}}^2}^2 \lesssim n^{-\frac{8\theta+2}{2+s}}\log\left(\frac{1}{\delta}\right),$$
where $\theta \in (-\frac{1}{4},0)$.
\end{theorem}

Theorem 2 shows that, when $n$ goes to infinity, the estimator $f_z$ can learn the conditional mean function $f^*$ by the rate ${\mathcal{O}}(n^{-\frac{8\theta+2}{2+s}})$ as the scale parameter $\sigma$ tends to 0 via ${\mathcal{O}}(n^{\theta})$  with $\theta \in (-\frac{1}{4},0)$, and the optimal rate is \begin{math} {\mathcal{O}}(n^{-1})\end{math} in the sense of approximation. Together with \cite{2015Learning}, where the scale parameter goes to infinity or is a constant large enough are discussed, this work completes the theoretical analysis for MCCR. And, the convergence rate is established when the noise is the mixture of symmetric distributions with some common conditions, which ensures the model to learn the unknown mean regression functions under mild conditions. Last, this result reveals that MCCR models hold favorable learning rate as well as adequate robustness via tending to 0 scale parameters.  All these lead to the merits of MCCR models in dealing with kinds of noises and outliers, some cases will be shown in Section 4.

\subsection{ Proofs of Theorem 1 and 2}
First, we display the following lemma which is needed to prove Theorem 2. It is given in \cite{2007Multi} and adopted in kinds of statistical learning works such as \cite{guo2013concentration, 2018Learning} and \cite{feng2020statistical}.

\begin{lemma}
Let $\mathcal{F}$ be a set of measurable functions on $\mathcal{Z}$, and $B,c>0$, $\tau \in [0,1]$ be constants such that each of $f\in \mathcal{F}$ satisfies $\Vert f\Vert_{\infty}\leq B$ and $\mathbb{E}f^2\leq c(\mathbb{E}f)^\tau$. If for some $a>0$ and $s\in (0,2)$,
$$\log \mathcal{N}_{2}(\mathcal{F},\eta)\leq a{\eta}^{-s},\forall \eta>0,$$
then there exists a constant $c_s$ such that for any $t>0$, with probability at least $1-e^{-t}$, for all $f\in \mathcal{F}$, there holds
$$\mathbb{E}f-\frac{1}{n}\sum_{i=1}^{n}f(z_i)\leq \frac{1}{2}{\zeta}^{1-\tau}(\mathbb{E}f)^{\tau}+c_{s}\zeta+2\left(\frac{ct}{n}\right)^{\frac{1}{2-\tau}}+\frac{18Bt}{n},$$
where
$$\zeta:=max\left\{c^{\frac{2-s}{4-2\tau+s\tau}}\left(\frac{a}{n}\right)^{\frac{2}{4-2\tau+s\tau}},
B^{\frac{2-s}{2+s}}\left(\frac{a}{n}\right)^{\frac{2}{2+s}}\right\}.$$
\end{lemma}
\
\\
\textbf{Proof of Theorem 1.} The techniques we used here is similar with \cite{fan2016consistency}. First, we prove the left inequality. From the definition of ${\varepsilon}^{\sigma}(f)$, we have
\[
\begin{split}
{\varepsilon}^{\sigma}(f)-{\varepsilon}^{\sigma}(f^*)&=
\int\limits_{\mathcal{X} \times \mathcal{Y}}l_{\sigma}(y-f(x))d\rho-\int\limits_{\mathcal{X} \times \mathcal{Y}}l_{\sigma}(y-f^*(x))d\rho  \\
&={\sigma}^2\int\limits_\mathcal{X}\int\limits_\mathcal{Y}
\left[-\exp\left(-\frac{(y-f(x))^2}{{\sigma}^2}\right)+\exp\left(-\frac{(y-f^*(x))^2}{{\sigma}^2}\right)\right]p_{y|x}(y)dyd{\rho}_\mathcal{X}(x)\\
&={\sigma}^2\int\limits_\mathcal{X}\int_{-\infty}^{+\infty}
\left[\exp\left(-\frac{t^2}{{\sigma}^2}\right)-\exp\left(-\frac{(t-(f(x)-f^*(x)))^2}{{\sigma}^2}\right)\right]p_{\varepsilon|x}(t)dtd{\rho}_\mathcal{X}(x)\\
\end{split}
\]
By Plancherel formula, we have
\[
\begin{split}
{\varepsilon}^{\sigma}(f)-{\varepsilon}^{\sigma}(f^*)&=
\frac{{\sigma}^3}{2\sqrt{\pi}}\int\limits_\mathcal{X}\int_{-\infty}^{+\infty}
\exp\left(-\frac{{\sigma}^2{\xi}^2}{4}\right)[1-\exp(-i{\xi}(f(x)-f^*(x)))]\widehat{p_{\varepsilon|x}}(\xi)d{\xi}d{\rho}_\mathcal{X}(x)\\
&=\frac{{\sigma}^3}{2\sqrt{\pi}}\int\limits_\mathcal{X}\int_{-\infty}^{+\infty}\exp\left(-\frac{{\sigma}^2{\xi}^2}{4}\right)
2{\sin}^2\left(\frac{\xi(f(x)-f^*(x))}{2}\right)\widehat{p_{\varepsilon|x}}(\xi)d{\xi}d{\rho}_\mathcal{X}(x)
\end{split}
\]
where $\widehat{p_{\varepsilon|x}}$ is the Fourier transform of $p_{\varepsilon|x}$, the second equation holds because ${\varepsilon}^{\sigma}(f)-{\varepsilon}^{\sigma}(f^*)$ is real for $\forall f\in \mathcal{H}$. From the linearity of the Fourier transformation and Assumption 1, we get
$$\widehat{p_{\varepsilon|x}}(\xi)=\sum_{i=1}^{K}w_{i}\widehat{p_{\varepsilon,i|x}}(\xi).$$
where $\widehat{p_{\varepsilon,i|x}}$ is the Fourier transformation of $p_{\varepsilon,i|x}$ ($i=1,2,\cdots,K$). Furthermore, from Assumption 2, it is known that exists positive constants $c_{0,i}, C_{0,i}$ with $i=1,2,\cdots,K$, such that
\[
\begin{split}
{\varepsilon}^{\sigma}(f)-{\varepsilon}^{\sigma}(f^*)
&=\frac{{\sigma}^3}{\sqrt{\pi}}\int\limits_\mathcal{X}\sum_{i=1}^{K}w_i\int_{-\infty}^{+\infty}\exp\left(-\frac{{\sigma}^2{\xi}^2}{4}\right)
{\sin}^2\left(\frac{\xi(f(x)-f^*(x))}{2}\right)\widehat{p_{\varepsilon,i|x}}(\xi)d{\xi}d{\rho}_\mathcal{X}(x)\\
&\geq\frac{{\sigma}^3}{\sqrt{\pi}}\int\limits_\mathcal{X}\sum_{i=1}^{K}w_i\int_{-c_{0,i}}^{c_{0,i}}\exp\left(-\frac{{\sigma}^2{\xi}^2}{4}\right)
{\sin}^2\left(\frac{\xi(f(x)-f^*(x))}{2}\right)C_{0,i}d{\xi}d{\rho}_\mathcal{X}(x)
\end{split}
\]
And, for $\forall x\in \mathcal{X}$, $|f(x)-f^*(x)| \leq 2M$. When $|\xi| \leq \frac{\pi}{2M}$, from Jordan's inequality, it holds that
$$\frac{{\xi}^{2}(f(x)-f^*(x))^2}{{\pi}^2} \leq {\sin}^2\left(\frac{\xi(f(x)-f^*(x))}{2}\right).$$
Then, let $c^{'}=\min\{c_{0,1},c_{0,2},\cdots,c_{0,K},\frac{\pi}{2M}\}$, so we have
\begin{equation}
\begin{split}
{\varepsilon}^{\sigma}(f)-{\varepsilon}^{\sigma}(f^*)& \geq
\frac{{\sigma}^3}{{\pi}^{\frac{5}{2}}}\int\limits_\mathcal{X}\sum_{i=1}^{K}w_i \int_{-c^{'}}^{c^{'}}
{\xi}^{2}\exp\left(-\frac{{\sigma}^2{\xi}^2}{4}\right)C_{0,i}(f(x)-f^*(x))^2d{\xi}d{\rho}_\mathcal{X}(x)\\
&:=c_{\sigma}\int\limits_\mathcal{X}(f(x)-f^*(x))^2d{\rho}_\mathcal{X}(x)
\end{split}
\end{equation}
where
\begin{equation}
\begin{split}
c_{\sigma}=\frac{{\sigma}^3}{{\pi}^{\frac{5}{2}}}\sum_{i=1}^{K}w_i C_{0,i}\int_{-c^{'}}^{c^{'}}{\xi}^2
\exp\left(-\frac{{\sigma}^2{\xi}^2}{4}\right)d{\xi}.
\end{split}
\end{equation}
is a positive constant. This implies that ${\varepsilon}^{\sigma}(f)\geq {\varepsilon}^{\sigma}(f^*)$ for any $f\in \mathcal{H}$. In other words,
$$f^*=\mathop{\arg\min}\limits_{f\in \mathcal{H}}{\varepsilon}^{\sigma}(f).$$
Second,
\[
\begin{aligned}
&{\varepsilon}^{\sigma}(f)-{\varepsilon}^{\sigma}(f^*)={\sigma}^2\int\limits_\mathcal{X}\int_{-\infty}^{+\infty} \left[\exp\left(-\frac{t^2}{{\sigma}^2}\right)-\exp\left(-\frac{(t-(f(x)-f^*(x)))^2}{{\sigma}^2}\right)\right]p_{\varepsilon|x}(t)dtd{\rho}_\mathcal{X}(x)\\
&:={\sigma}^2\int\limits_\mathcal{X}[F_{x}(f(x)-f^*(x))-F_{x}(0)]d{\rho}_\mathcal{X}(x)
\end{aligned}
\]
where $F_{x}:\mathbb{R} \to \mathbb{R}$ is defined as
$$F_{x}(u):=1-\int_{-\infty}^{+\infty}\exp\left(-\frac{(t-u)^2}{{\sigma}^2}\right)p_{\varepsilon|x}(t)dt,x\in \mathcal{X}.$$
From Taylor expansions, we know that
$$F_{x}(f(x)-f^*(x))-F_{x}(0)=F_{x}^{'}(0)(f(x)-f^*(x))+\frac{F_{x}^{''}({\zeta}_x)}{2}(f(x)-f^*(x))^2,$$
where for $\forall x\in \mathcal{X}$, ${\zeta}_{x}$ is between 0 and $f(x)-f^*(x)$. Due to the noise is symmetric with 0 as the center, for $\forall x\in \mathcal{X}$, we have
$$F_{x}^{'}(0)=-\int_{-\infty}^{+\infty}\exp\left(-\frac{t^2}{{\sigma}^2}\right)\left(\frac{2t}{{\sigma}^2}\right)p_{\varepsilon|x}(t)dt=0.$$
In addition, for $\forall x\in \mathcal{X}$,
\[
\begin{split}
F_{x}^{''}({\zeta}_x)&=\int_{-\infty}^{+\infty}2\exp\left(-\frac{(t-{\zeta}_x)^2}{{\sigma}^2}\right)\left(\frac{{\sigma}^2-2(t-{\zeta}_x)^2}{{\sigma}^4}\right)p_{\varepsilon|x}(t)dt \leq \frac{2}{{\sigma}^2}.
\end{split}
\]
Therefore,
\begin{equation}
\begin{split}
{\varepsilon}^{\sigma}(f)-{\varepsilon}^{\sigma}(f^*)&={\sigma}^2\int\limits_\mathcal{X}0+\frac{F_{x}^{''}({\zeta}_x)}{2}(f(x)-f^*(x))^2d{\rho}_\mathcal{X}(x)\\
&\leq \int\limits_\mathcal{X}(f(x)-f^*(x))^2d{\rho}_\mathcal{X}(x)=\Vert f-f^*\Vert_{L_{\rho_\mathcal{X}}^2}^2.
\end{split}
\end{equation}
Combine (4) and (6), we obtain
$$c_{\sigma}\Vert f-f^*\Vert_{L_{\rho_\mathcal{X}}^2}^2\leq {\varepsilon}^{\sigma}(f)-{\varepsilon}^{\sigma}(f^*)\leq \Vert f-f^*\Vert_{L_{\rho_\mathcal{X}}^2}^2,$$
where $c_{\sigma}$ is a positive constant given in (5). This completes the proof of Theorem 1.

\begin{remark}
When $\sigma$ changes with the sample size $n$, we will make further analysis here. From (5), it is known that there exists a constant $A_i\in [-c^{'},c^{'}], i=1,2,\cdots,K$, such that,
\[
\begin{split}
c_{\sigma}&=\frac{{\sigma}^3}{{\pi}^{\frac{5}{2}}}\sum_{i=1}^{K}w_i{A_i}^2
\exp\left(-\frac{{\sigma}^2{A_i}^2}{4}\right)C_{0,i}(c^{'}+c^{'})\\
&={\sigma}^3\sum_{i=1}^{K}\exp\left(-\frac{{\sigma}^2{A_i}^2}{4}\right)2{\pi}^{-\frac{5}{2}}c^{'}w_i{A_i}^2C_{0,i}\\
&:={\sigma}^3\sum_{i=1}^{K}\exp\left(-\frac{{\sigma}^2{A_i}^2}{4}\right)c_1,
\end{split}
\]
where
$$c_1=2{\pi}^{-\frac{5}{2}}c^{'}w_i{A_i}^2C_{0,i},$$
is a positive constant.
\end{remark}
\
\\ \textbf{Proof of Theorem 2.}
First, we prove that Theorem 2 satisfies the conditions of Lemma 1. We assume that the definition of the function set $\mathcal{F}_{\mathcal{H}}$ is as follows
$$\mathcal{F}_{\mathcal{H}}=\left\{g|g(z)=-{\sigma}^2\exp\left\{-\frac{(y-f(x))^2}{{\sigma}^2}\right\}+{\sigma}^2\exp\left\{-\frac{(y-f^*(x))^2}{{\sigma}^2}\right\}
,f\in \mathcal{H},z\in \mathcal{Z}\right\}.$$
Then, for $\forall g\in \mathcal{F}_{\mathcal{H}}$, we have
$$\Vert g\Vert_{\infty}\leq {\sigma}^2+{\sigma}^2\leq 2{\sigma}^2,$$
and
$$\mathbb{E}g={\varepsilon}^{\sigma}(f)-{\varepsilon}^{\sigma}(f^*).$$
Moreover, introducing the auxiliary function $h(t)=-{\sigma}^2\exp\{-\frac{t^2}{\sigma^2}\}$, $t \in \mathbb{R}$, it is easy to see that $\| h^{'} \|_{\infty}=\sqrt{2/e} \sigma$. By taking $t_1=y-f(x)$, $t_2=y-f^{*}(x)$ and applying the mean value theorem to $h$, we see that
\[
\begin{split}
\mathbb{E}g^2&=\int\limits_\mathcal{Z}\left(-{\sigma}^2\exp\left\{-\frac{(y-f(x))^2}{{\sigma}^2}\right\}+{\sigma}^2\exp\left\{-\frac{(y-f^*(x))^2}{{\sigma}^2}\right\}\right)^2d{\rho}\\
&\leq \int\limits_\mathcal{Z}\|h^{'} \|_{\infty}^2(f(x)-f^*(x))^2d{\rho}\\
&\leq 2e^{-1}{\sigma}^2\int\limits_\mathcal{Z}(f(x)-f^*(x))^2d{\rho}\\
&\lesssim {\sigma}^{-1}\left(\sum_{i=1}^{K}\exp\left\{-\frac{\sigma^2A_i^2}{4}\right\}\right)^{-1}\mathbb{E}g.
\end{split}
\]
On the other hand, for $\forall g_1,g_2\in \mathcal{F}_{\mathcal{H}},$ $\exists f_1,f_2\in \mathcal{H}$, such that
$$g_1(z)=-{\sigma}^2\exp\left\{-\frac{(y-f_1(x))^2}{{\sigma}^2}\right\}+{\sigma}^2\exp\left\{-\frac{(y-f^*(x))^2}{{\sigma}^2}\right\},$$
and
$$g_2(z)=-{\sigma}^2\exp\left\{-\frac{(y-f_2(x))^2}{{\sigma}^2}\right\}+{\sigma}^2\exp\left\{-\frac{(y-f^*(x))^2}{{\sigma}^2}\right\}.$$
So, we have
$$\| g_1-g_2 \|_{\infty} \leq \sqrt{2/e} \sigma \cdot \| f_1-f_2 \|_{\infty}.$$
Then, under the assumption of complexity and $0<s<2$, the $l^{2}$-empirical covering numbers of $\mathcal{F}_{\mathcal{H}}$ and $\mathcal{H}$ have the following relation
$$\log \mathcal{N}_2(\mathcal{F}_{\mathcal{H}},\eta)\leq \log \mathcal{N}_2(\mathcal{H},\frac{\eta}{\sqrt{2/e}\sigma})\lesssim \sigma^s \eta^{-s}.$$
Now, applying Lemma 1 to the random variable $g$ with $B=2\sigma^2$, $c={\sigma}^{-1}\left(\sum_{i=1}^{K}\exp\left\{-\frac{\sigma^2A_i^2}{4}\right\}\right)^{-1}$, $\tau=1$, $a=\sigma^s$, then for any $0<{\delta}<1$, with probability at least $1-\delta$, it holds that
$$[{\varepsilon}^{\sigma}(f)-{\varepsilon}^{\sigma}(f^*)]-[{\varepsilon}_{z}^{\sigma}(f)-{\varepsilon}_{z}^{\sigma}(f^*)]-\frac{1}{2}[{\varepsilon}^{\sigma}(f)-{\varepsilon}^{\sigma}(f^*)]
\lesssim (\sigma^{\frac{3s-2}{2+s}}n^{-\frac{2}{2+s}}+\sigma^{-1}n^{-1})\log\left(\frac{1}{\delta}\right).$$
Because $f_z=\mathop{\arg\min}\limits_{f\in \mathcal{H}}{\varepsilon}^{\sigma}_{z}(f)$, we have
\[
\begin{split}
{\varepsilon}^{\sigma}(f_z)-{\varepsilon}^{\sigma}(f^*) & \leq 2[{\varepsilon}^{\sigma}(f_z)-{\varepsilon}^{\sigma}(f^*)]-2[{\varepsilon}_{z}^{\sigma}(f_z)-{\varepsilon}_{z}^{\sigma}(f^*)]-[{\varepsilon}^{\sigma}(f_z)-{\varepsilon}^{\sigma}(f^*)]\\
& \lesssim (\sigma^{\frac{3s-2}{2+s}}n^{-\frac{2}{2+s}}+\sigma^{-1}n^{-1})\log\left(\frac{1}{\delta}\right).
\end{split}
\]
Therefore, according to Theorem 1, for any $0<{\delta}<1$, with probability at least $1-\delta$, it holds that
\[
\begin{split}
\Vert f_{z}-f^*\Vert_{L_{\rho_\mathcal{X}}^2}^2 & \leq \frac{1}{c_\sigma}[{\varepsilon}^{\sigma}(f_z)-{\varepsilon}^{\sigma}(f^*)]\\
& \lesssim (\sigma^{\frac{3s-2}{2+s}}n^{-\frac{2}{2+s}}+\sigma^{-1}n^{-1})\sigma^{-3}\left(\sum_{i=1}^{K}\exp\left\{-\frac{\sigma^2A_i^2}{4}\right\}\right)^{-1}\log\left(\frac{1}{\delta}\right)\\
& \lesssim (\sigma^{-\frac{8}{2+s}}n^{-\frac{2}{2+s}}+\sigma^{-4}n^{-1})\log\left(\frac{1}{\delta}\right).
\end{split}
\]
Let $\sigma=n^{\theta}$ with $\theta \in (-\frac{1}{4},0)$, then we have
$$\Vert f_{z}-f^*\Vert_{L_{\rho_\mathcal{X}}^2}^2 \lesssim n^{-\frac{8\theta+2}{2+s}}\log\left(\frac{1}{\delta}\right).$$
This completes the proof of Theorem 2.

\vspace*{+0.1cm}
\section{Simulations }
In this section, we use synthetic data of finite samples to show the effectiveness of MCCR model under mixed symmetric noises and compare the robustness of MCCR to Huber and least squares (LS) regression models. Assume further that the space $\mathcal{H}$ is a bounded subset of a reproducing kernel Hilbert space $\mathcal{H}_{\mathcal{K}}$, where $\mathcal{K}$ is a Mercer kernel. At this time, the solution (3) of the MCCR model is equivalent to
\begin{equation}
f_z=\mathop{\arg\min}\limits_{f\in \mathcal{H}_{\mathcal{K}}}\frac{1}{n}{\sum_{i=1}^{n}l_\sigma(y_i-f(x_i))}+\lambda\Vert f \Vert_{\mathcal{K}}^2,
\end{equation}
where $\lambda$ is a positive regularization parameter. The representor theorem ensures that $f_z$ can be modeled by
$$f_{z}(x)=\sum_{i=1}^{n}\alpha_{i}\mathcal{K}(x,x_i)+b_0,$$
where $\bm{\alpha}=(\alpha_1,\cdots,\alpha_n)^\top \in \mathbb{R}^n$ and $b_0 \in \mathbb{R}$ are learned from (7). We use Gaussian kernel $\mathcal{K}(x,x')=\exp\{-\frac{\Vert x-x'\Vert^2}{h^2}\}$ and the iterative weighted least squares to perform the algorithms. The initial values of the iterations  for MCCR regression and Huber regression are $\bm{\alpha}=(0,\cdots,0)^\top$ and $b_0=0$. The scale parameter $\sigma$,  bandwidth parameter $h$ and regularization parameter $\lambda$ in $l_{\sigma}$ loss and Huber's loss, the bandwidth parameter $h$ and  regularization parameter $\lambda$ in LS regression are all selected by the rule of five-fold cross validation.

\vspace*{+0.2cm}
\noindent \textbf{Example 1.} The real function is $f(x)=e^{-7.5x}\cos(10\pi x)$, where $X_{i}$ are independent sample points from a uniform distribution $U(0,1)$, and $Y_{i}=f(X_{i})+\varepsilon_{i}$ with $\varepsilon_{i}\sim 0.8N(0,0.1^2)+0.2N(0,0.5^2)$. $N(0,0.1^2)$ is the background noise and $N(0,0.5^2)$ is used to generate outliers, $i=1, 2, \cdots, n$ and $n=200$. The estimates  are shown in Fig.1.\\

\begin{figure*}[htbp]
\centering
\subfigure[]{
\includegraphics[width=2.5in,height=2.5in]{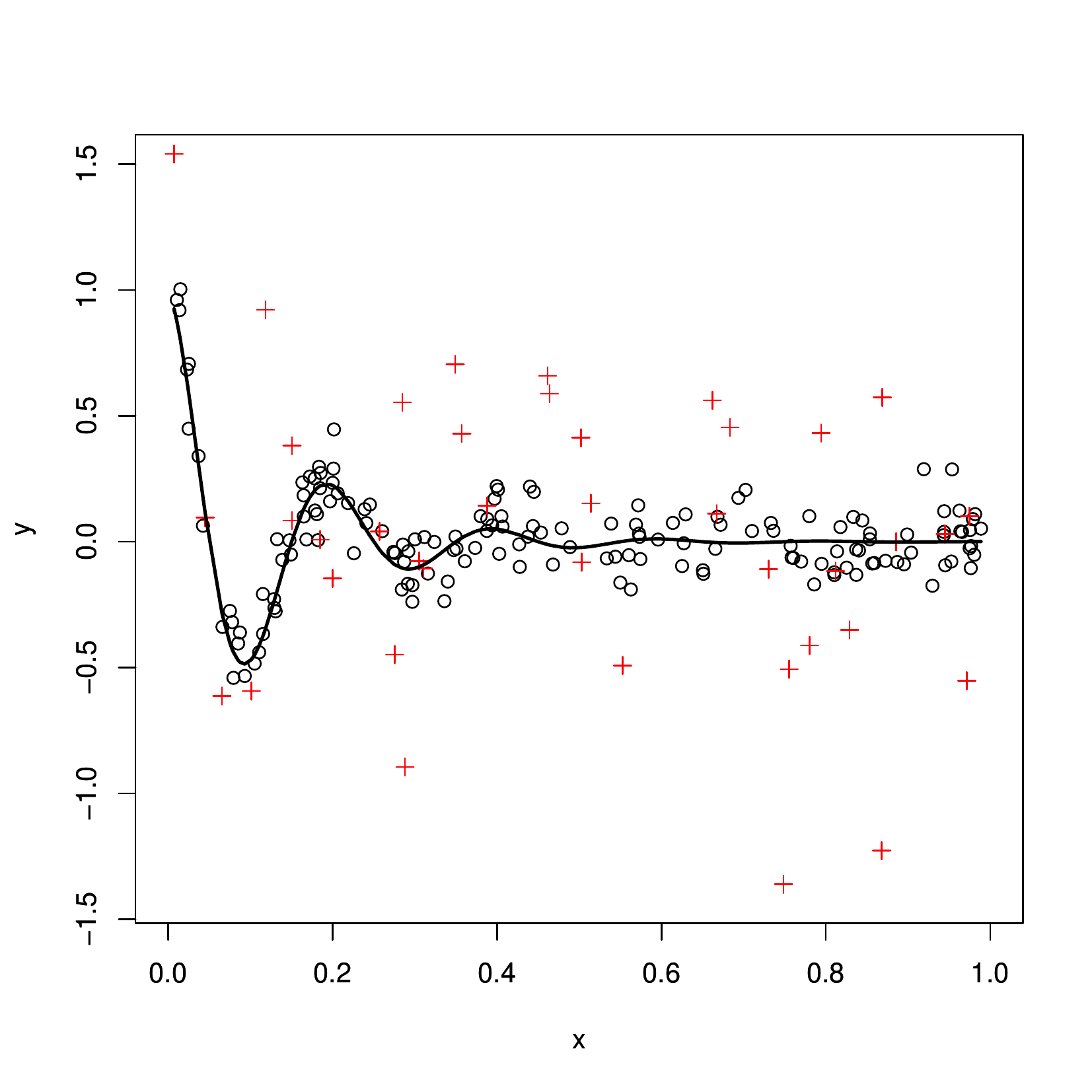}}
\hspace*{1cm}
\subfigure[]{
\includegraphics[width=2.5in,height=2.5in]{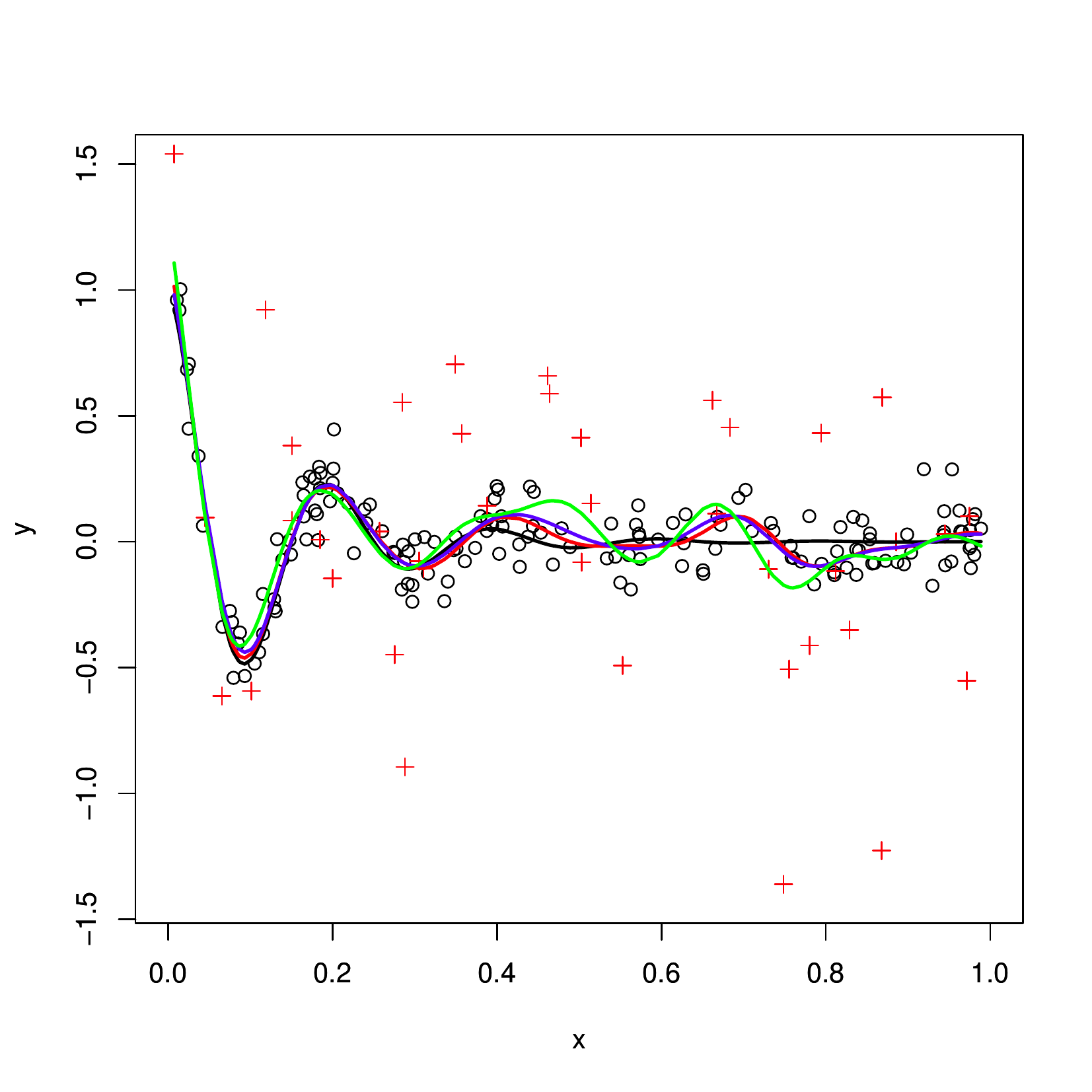}}
\caption{Real function, estimates and scatter plots of Example 1. (a) Real function (black solid curve) and scatter plots. Circles are sample points with background noise and red crosses are outliers. (b) Red, blue and green solid curves are the estimates of MCCR,  Huber and LS regressions, respectively.}
\label{fig:1}       
\end{figure*}

\vspace*{+0.2cm}
\noindent \textbf{Example 2.} The real function is $f(x)= -1+1.5x+0.2\phi(x-0.6)$, where $\phi(x)$ is the density function of normal distribution $N(0,0.04^2)$, $X_{i}$ are independent sample points from a uniform distribution $U(0,1)$, and $Y_{i}=f(X_{i})+\varepsilon_{i}$ with $\varepsilon_{i}\sim 0.8N(0,0.1^2)+0.2N(0,1)$. $N(0,0.1^2)$ is the background noises and $N(0,1)$ is used to generate outliers, $i=1, 2, \cdots, n$ and $n=200$. The estimates  are shown in Fig.2.

\begin{figure*}[htbp]
\centering
\subfigure[]{
\includegraphics[width=2.5in,height=2.5in]{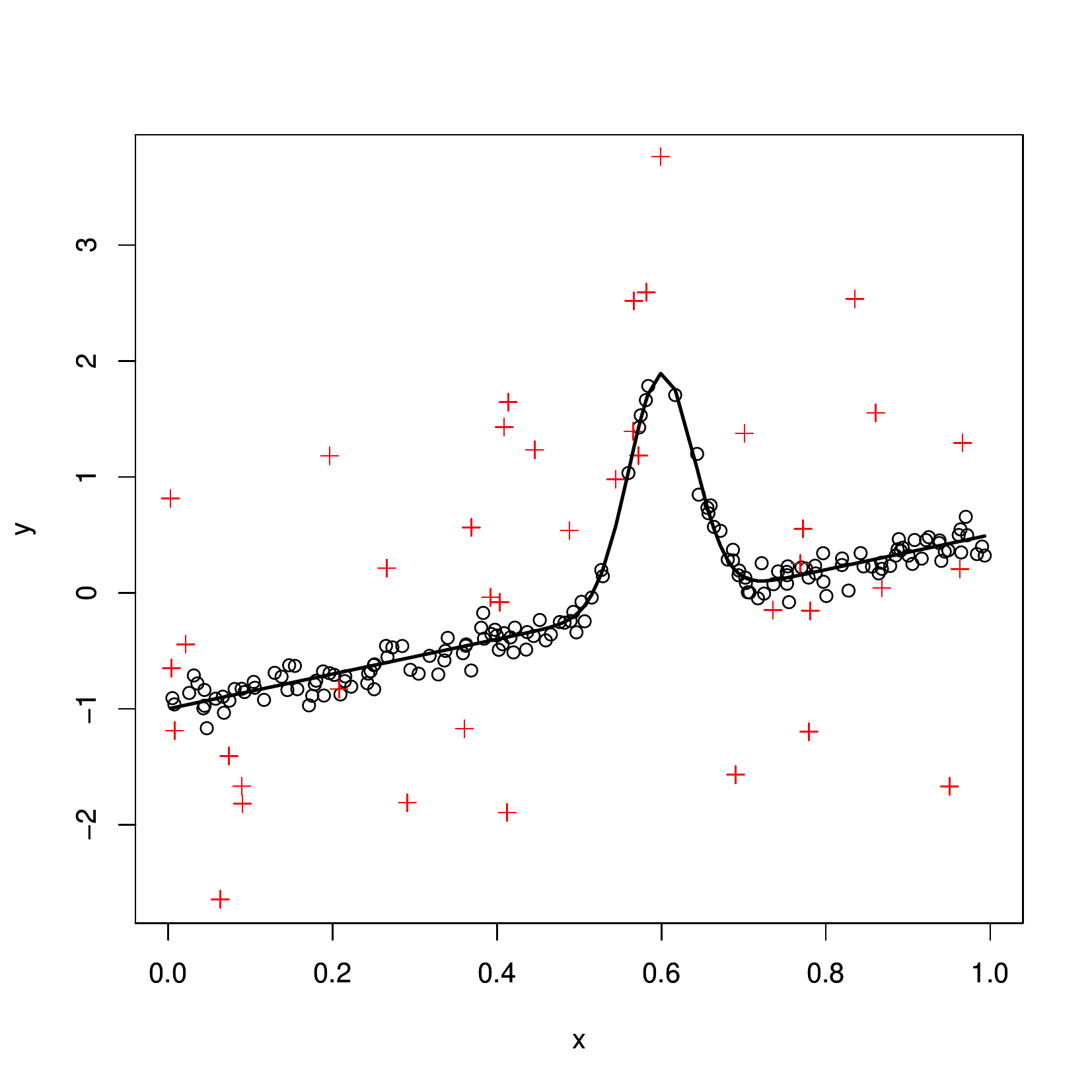}}
\hspace*{1cm}
\subfigure[]{
\includegraphics[width=2.5in,height=2.5in]{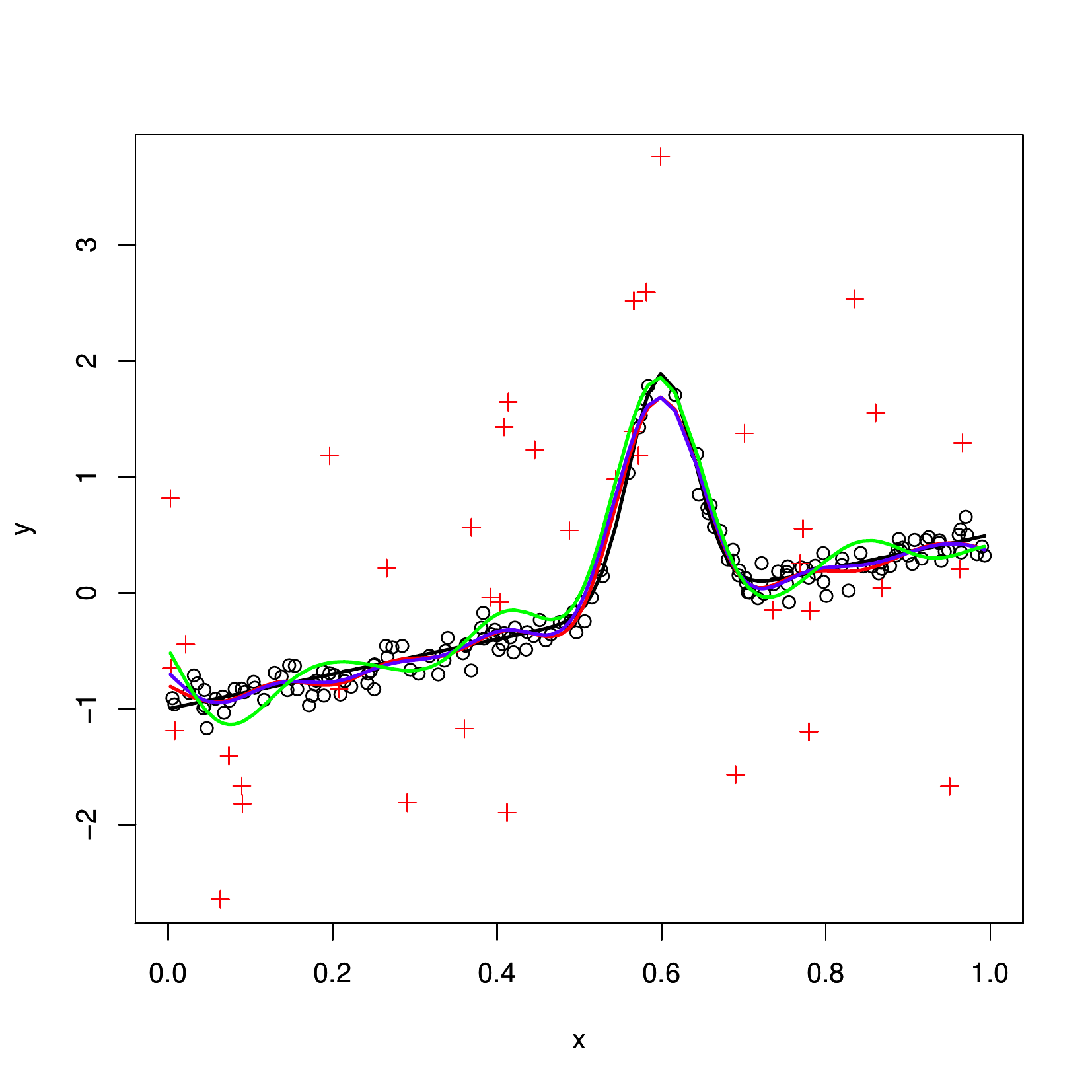}}
\caption{Real function, estimates and scatter plots of Example 2. (a) Real function (black solid curve) and scatter plots. Circles are sample points with background noise and red crosses are outliers. (b) Red, blue and green solid curves are the estimates of  MCCR,  Huber  and LS  regressions, respectively.}
\label{fig:2}
\end{figure*}

\vspace*{+0.2cm}
\noindent \textbf{Example 3.} The real function is $f(x)=\frac{\sin(x)}{x}$, where $X_{i}$ are independent sample points from a uniform distribution $U(-10,10)$, and $Y_{i}=f(X_{i})+\varepsilon_{i}$ with $\varepsilon_{i}\sim 0.8N(0,0.1^2)+0.2Cauchy(0,0.2)$. $N(0,0.1^2)$ is the background noise and $Cauchy(0,0.2)$ is used to generate outliers, $i=1, 2, \cdots, n$ and $n=200$. The estimates are shown in Fig.3.

\begin{figure*}[htbp]
\centering
\subfigure[]{
\includegraphics[width=2.5in,height=2.5in]{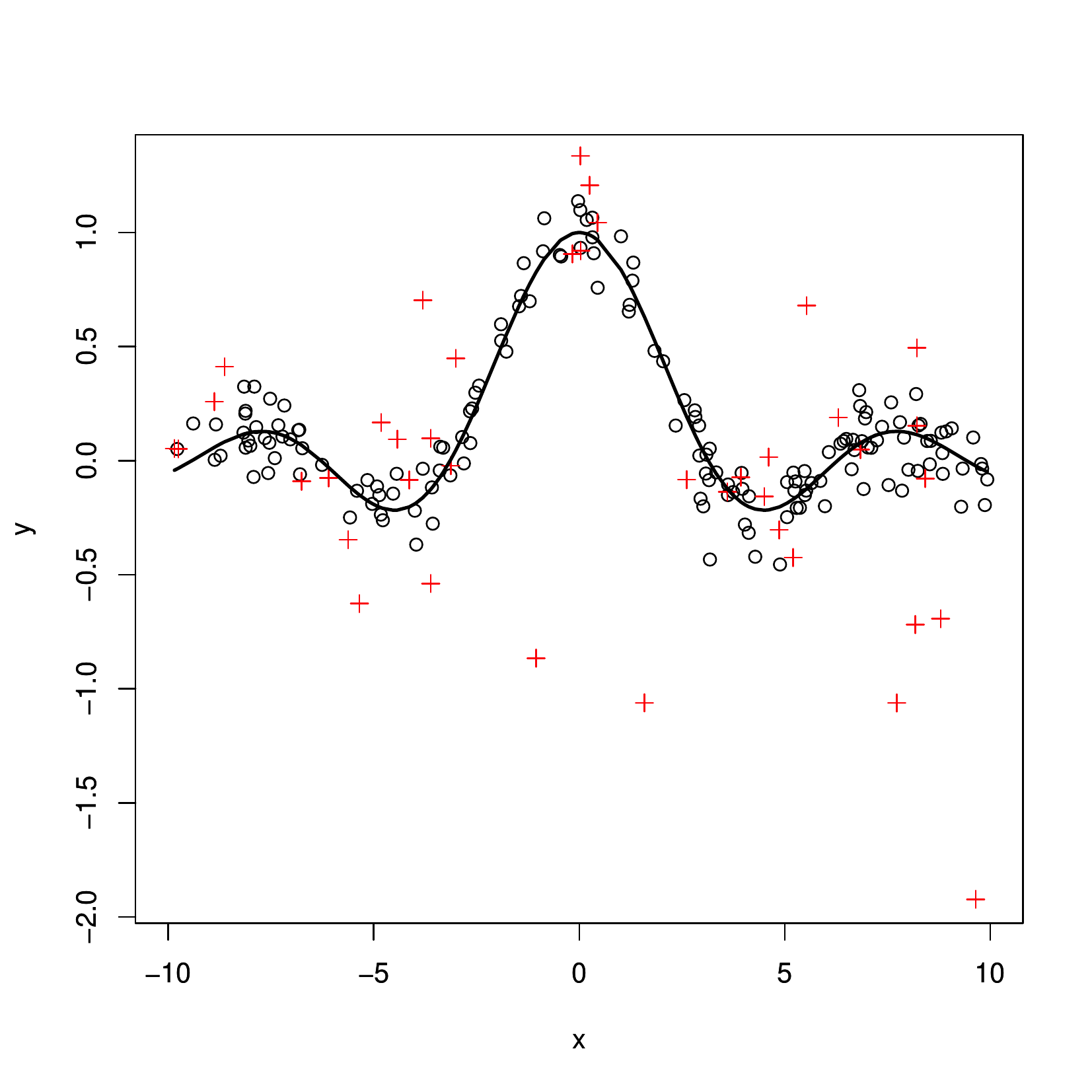}}
\hspace*{1cm}
\subfigure[]{
\includegraphics[width=2.5in,height=2.5in]{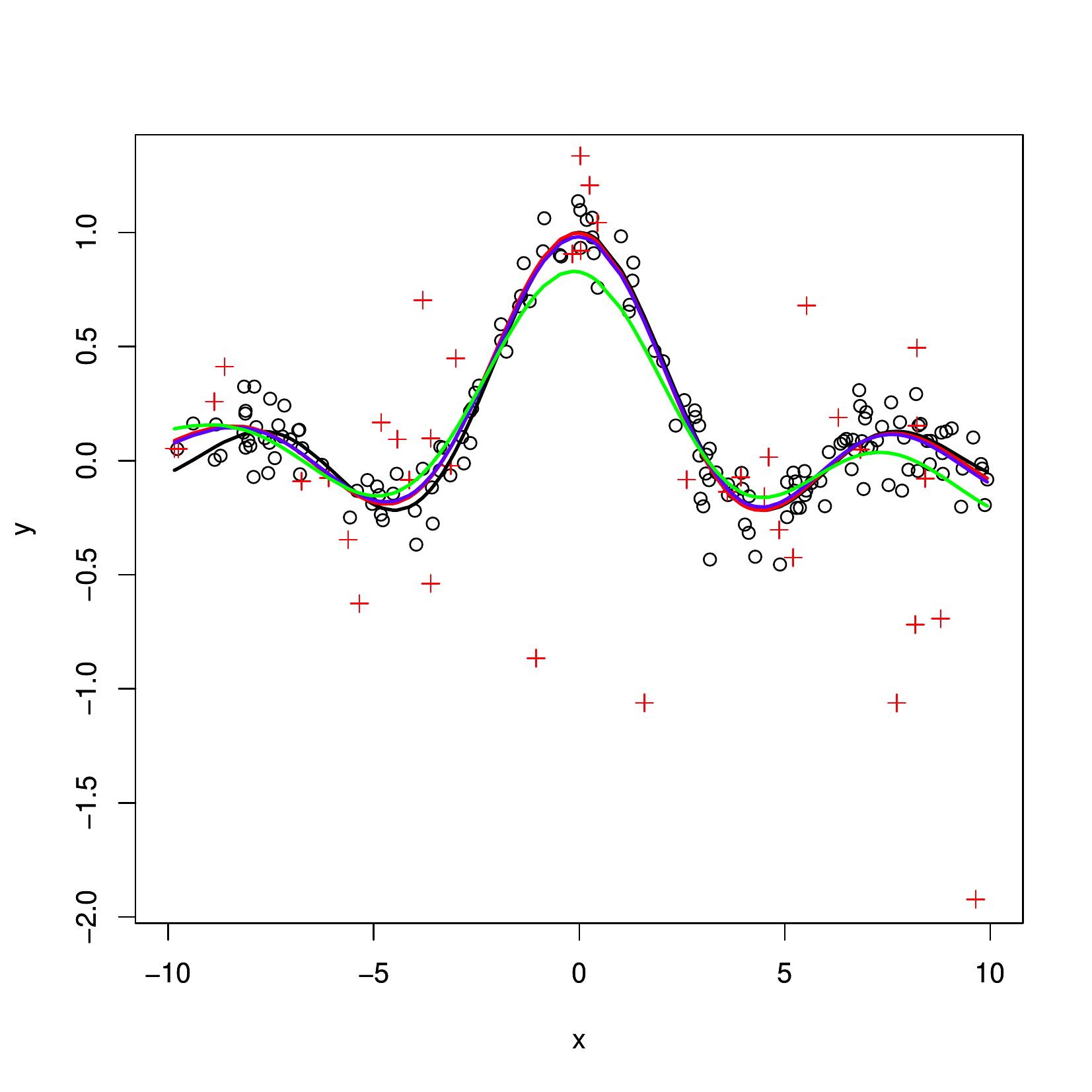}}
\caption{Real function, estimates and scatter plots of Example 3. (a) Real function (black solid curve) and scatter plots. Circles are sample points with background noise and red crosses are outliers. (b) Red, blue and green solid curves are the estimates of MCCR,  Huber and LS regressions, respectively.}
\label{fig:3}
\end{figure*}

Estimates shown in these three Figures tell us that the MCCR has a comparable performance with Huber regression on fitness and robustness, and both of them outperform LS regression when data contains outliers. Moreover, for each example, the simulations are repeated 100 times and the mean square error(MSE) between the estimator $f_z$ and the real function $f^*$ are given in Table 1 ( in brackets are the standard deviations), which supports the similar conclusion. Furthermore, let $\theta=-0.2$ and the sample size  $n$ goes from $100$ to $200, 300, \cdots, 600$, MSEs and their standard deviations are shown in Table 2 and Fig.4, which provide some evidence to Theorem 2.

\begin{table}[htbp]
\centering
\caption{The MSE between $f_z$ and $f^*$ of MCCR, Huber, and LS regression}
\label{tab:1}       
\begin{tabular}{llll}
\hline\noalign{\smallskip}
  & MCCR & Huber & LS  \\
\noalign{\smallskip}\hline\noalign{\smallskip}
Example 1 & $\mathbf{0.0014}(0.0006)$ & 0.0017(0.0006) & 0.0046(0.0015)\\
Example 2 & $\mathbf{0.0027}(0.0008)$ & 0.0045(0.0018) & 0.0165(0.0062)\\
Example 3 & $\mathbf{0.0006}(0.0004)$ & 0.0007(0.0004) & 0.2595(1.5240)\\
\noalign{\smallskip}\hline
\end{tabular}
\end{table}

\begin{table}[htbp]
\centering
\caption{The MSE of MCCR for sample size $n$}
\label{tab:2}       
\begin{tabular}{llll}
\hline\noalign{\smallskip}
  &Example 1 & Example 2 & Example 3 \\
\noalign{\smallskip}\hline\noalign{\smallskip}
 n=100 & 0.0034(0.0025) & 0.0066(0.0275) & 0.0013(0.0006) \\
 n=200 & 0.0017(0.0025) & 0.0035(0.0081) & 0.0006(0.0003) \\
 n=300 & 0.0010(0.0003) & 0.0031(0.0095) & 0.0004(0.0002) \\
 n=400 & 0.0007(0.0002) & 0.0026(0.0078) & 0.0003(0.0001) \\
 n=500 & 0.0006(0.0002) & 0.0013(0.0002) & 0.0003(0.0001) \\
 n=600 & 0.0005(0.0001) & 0.0011(0.0002) & 0.0002(0.0001) \\
\noalign{\smallskip}\hline
\end{tabular}
\end{table}

\begin{figure}[htbp]
\centering
  \includegraphics[width=10cm,height=7cm]{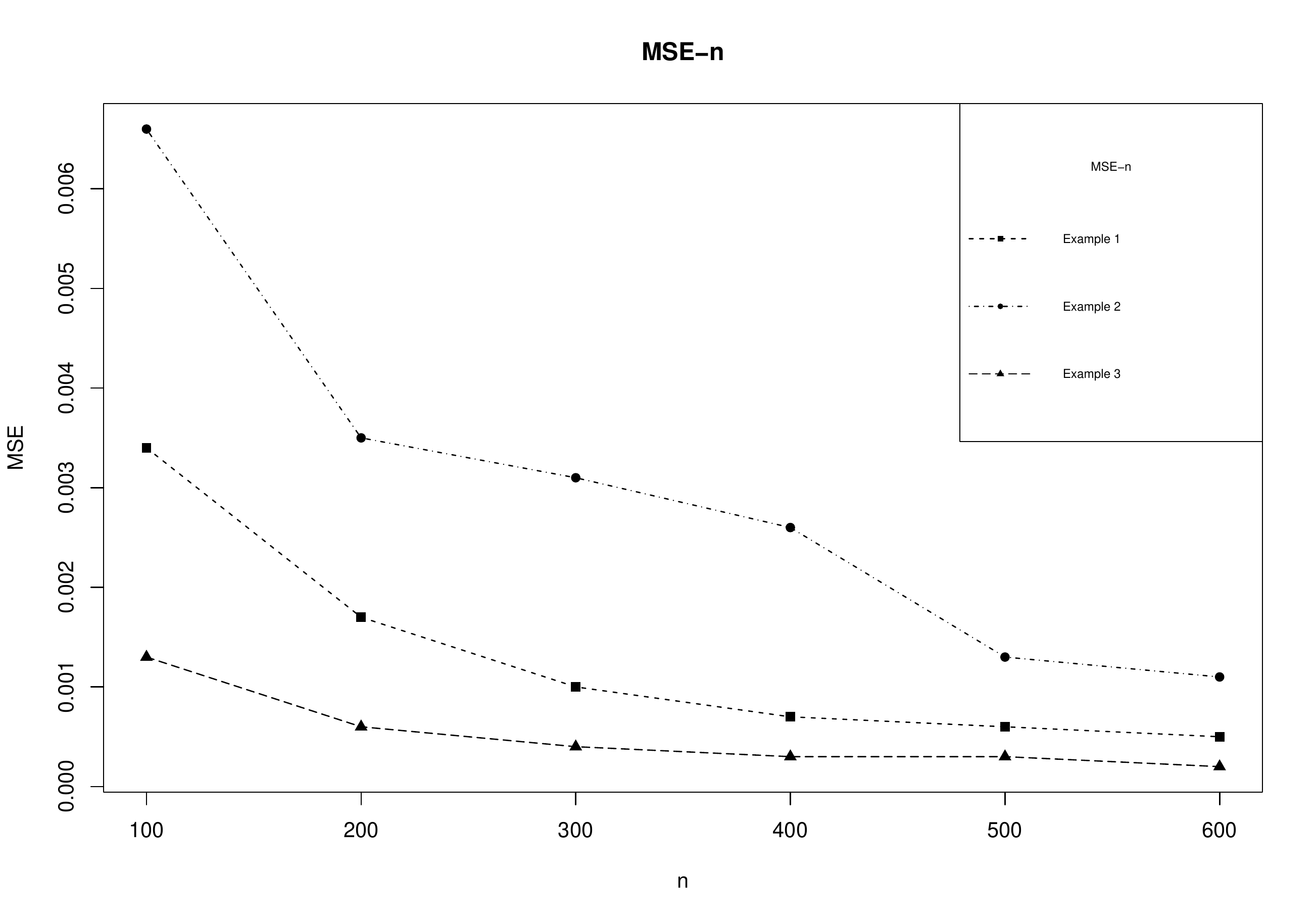}
\caption{MSE of MCCR  for  sample size $n$ .}
\label{fig:4}       
\end{figure}

\vspace*{+1cm}
\section{Application}
In this section, we use MCCR, Huber and LS regression to learn from the speed-flow data, which was firstly studied in \cite{petty1996freeway} and is publicly available from the \texttt{R}-package \texttt{hdrcde}. Let $X$ be the traffic flow, evaluated by the number of vehicles per hour per lane,  $Y$ be the speed of the vehicle, measured in miles per hour, scatter plots of two data sets collected on two separate lanes (lane 2 and lane 3) of the 4-lane Californian highway I-880 in 1993 are displayed in Fig.5. Red, blue and green solid curves are the estimates of MCCR, Huber and LS regressions, respectively. The algorithm is the iterative weighted least squares and the hyperparameter selection method is five-fold cross-validation. The results show that MCCR holds the best robustness, LS regression is the worst and Huber regression is between the two.

\begin{figure*}
\subfigure[]{
\includegraphics[width=2.5in,height=2.5in]{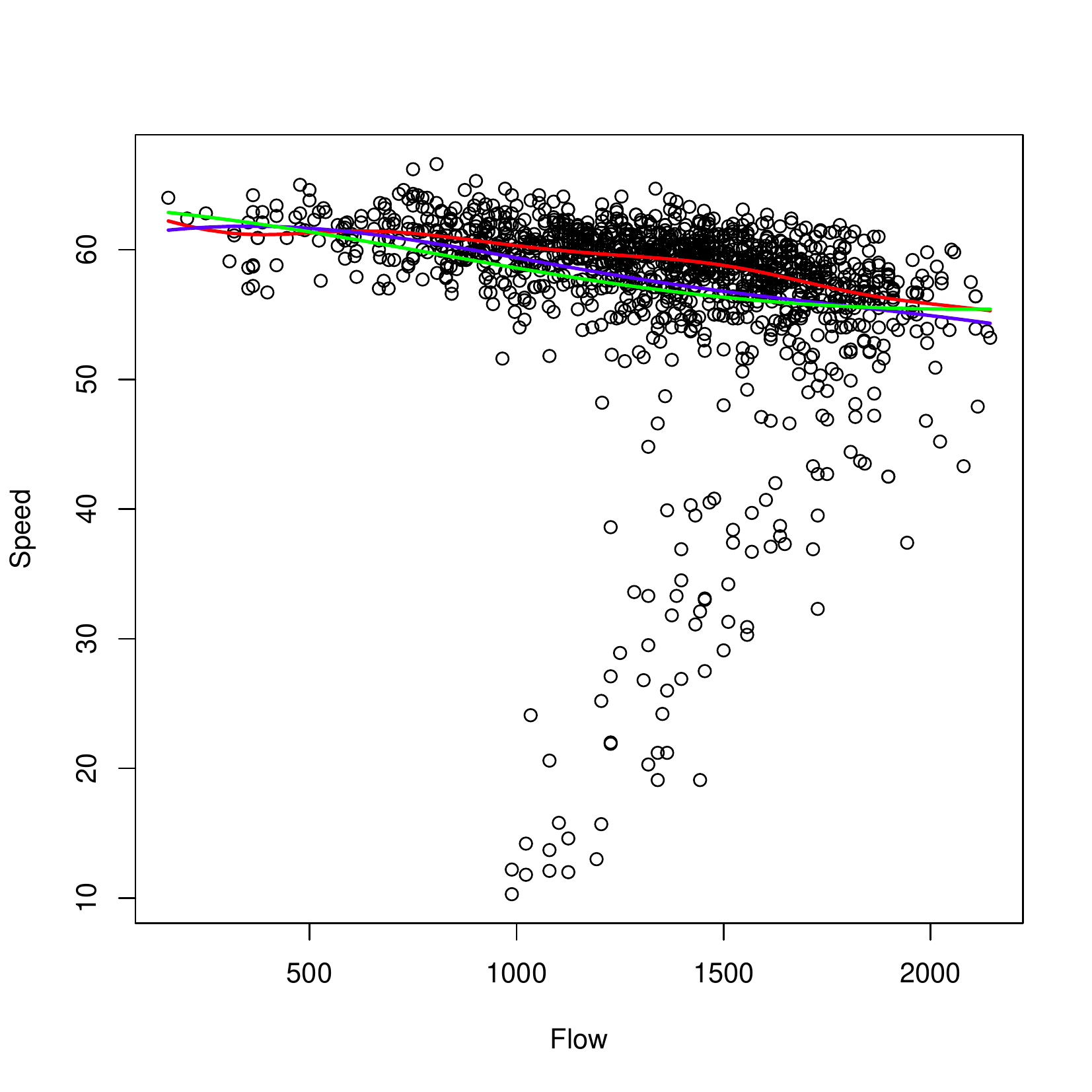}}
\hspace*{1cm}
\subfigure[]{
\includegraphics[width=2.5in,height=2.5in]{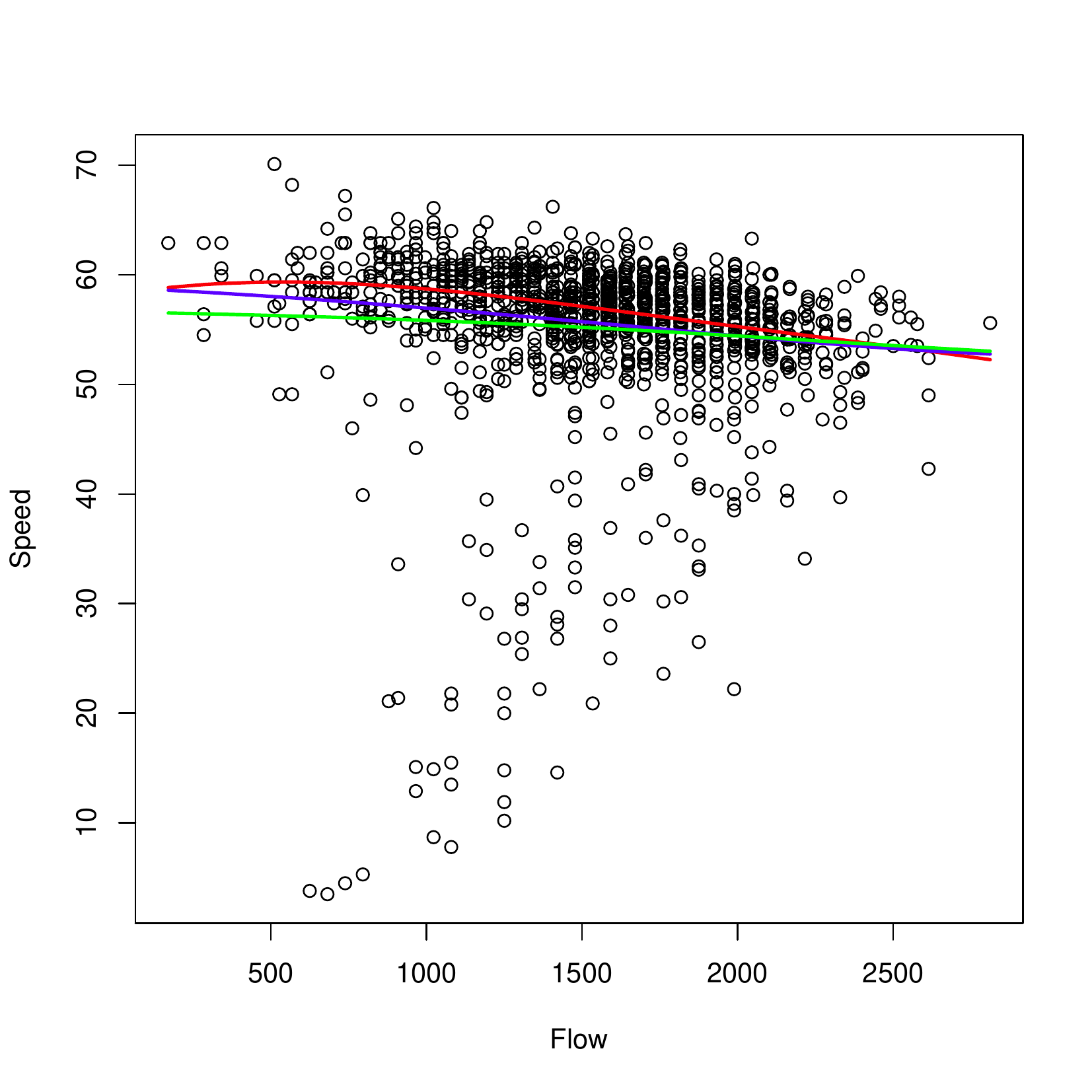}}
\caption{Estimates and scatter plots of  the speed-flow data. Red, blue and green solid curves are the estimates of  MCCR,  Huber  and LS  regressions, respectively. (a) Speed-Flow data for lane $2$ ; (b) Speed-Flow data for lane 3.}
\label{fig:5}       
\end{figure*}

\section{Conclusion}
Maximum Correntropy Criterion Regression models with tending-to-zero scale parameters are theoretically studied within the frame of statistical learning. The model can learn the conditional mean function $f^*$ with the rate ${\mathcal{O}}(n^{-\frac{8\theta+2}{2+s}})$ as the scale parameter $\sigma$ tends to 0 via ${\mathcal{O}}(n^{\theta})$, where $\theta \in (-\frac{1}{4},0)$ and $s$ is the complexity parameter of the hypothesis space. The optimal convergence rate of estimator is \begin{math} {\mathcal{O}}(n^{-1})\end{math} in the sense of approximation. This study implies that MCCR models can attain both accurate approximation and adequate robustness when the sample size is large enough.  Simulations and applications of MCCR, Huber and LS regression models are evaluated, both of them show some evidences for that the MCCR model holds the best robustness.\par
\ \\
\ \\
\textbf{Acknowledgements}
\ \\
This research was supported by the National Science Foundation of China (Grant NO. 11671012), the Key University Science Research Project of Anhui, China (Grant NO. KJ2017A028) and  the Open Project Program of School of Mathematical Sciences of Anhui University (Grant NO. Y01002431).
\ \\
\ \\


\bibliographystyle{apalike}
\bibliography{1}

\begin{thebibliography}{}

\bibitem[Anthony and Bartlett, 1999]{anthony2009neural}
Anthony, M. and Bartlett, P.~L. (1999).
\newblock {\em Neural Network Learning: Theoretical Foundations}.
\newblock Cambridge University Press, Cambridge.

\bibitem[Chen and Pr{\'\i}ncipe, 2012]{2012Maximum}
Chen, B. and Pr{\'\i}ncipe, J.~C. (2012).
\newblock Maximum correntropy estimation is a smoothed map estimation.
\newblock {\em IEEE Signal Processing Letters}, 19(8):491--494.

\bibitem[Cucker and Zhou, 2007]{cucker2007learning}
Cucker, F. and Zhou, D.~X. (2007).
\newblock {\em Learning Theory: An Approximation Theory Viewpoint}.
\newblock Cambridge University Press, Cambridge.

\bibitem[Fama and Roll, 1968]{fama1968some}
Fama, E.~F. and Roll, R. (1968).
\newblock Some properties of symmetric stable distributions.
\newblock {\em Journal of the American Statistical Association},
  63(323):817--836.

\bibitem[Fan et~al., 2016]{fan2016consistency}
Fan, J., Hu, T., Wu, Q., and Zhou, D.~X. (2016).
\newblock Consistency analysis of an empirical minimum error entropy algorithm.
\newblock {\em Applied and Computational Harmonic Analysis}, 41(1):164--189.

\bibitem[Feng et~al., 2020]{feng2020statistical}
Feng, Y., Fan, J., and Suykens, J. A.~K. (2020).
\newblock A statistical learning approach to modal regression.
\newblock {\em Journal of Machine Learning Research}, 21(2):1--35.

\bibitem[Feng et~al., 2015]{2015Learning}
Feng, Y., Huang, X., Shi, L., Yang, Y., and Suykens, J. A.~K. (2015).
\newblock Learning with the maximum correntropy criterion induced losses for
  regression.
\newblock {\em Journal of Machine Learning Research}, 16:993--1034.

\bibitem[Feng and Wu, 2020]{feng2020learning}
Feng, Y. and Wu, Q. (2020).
\newblock Learning under $(1+\varepsilon)$-moment conditions.
\newblock {\em Applied and Computational Harmonic Analysis}, 49(2):495--520.

\bibitem[Feng and Ying, 2020]{2018Learning}
Feng, Y. and Ying, Y. (2020).
\newblock Learning with correntropy-induced losses for regression with mixture
  of symmetric stable noise.
\newblock {\em Applied and Computational Harmonic Analysis}, 48(2):795--810.

\bibitem[Gunduz and Principe, 2009]{gunduz2009correntropy}
Gunduz, A. and Principe, J.~C. (2009).
\newblock Correntropy as a novel measure for nonlinearity tests.
\newblock {\em Signal Processing}, 89(1):14--23.

\bibitem[Guo and Zhou, 2013]{guo2013concentration}
Guo, Z.~C. and Zhou, D.~X. (2013).
\newblock Concentration estimates for learning with unbounded sampling.
\newblock {\em Advances in Computational Mathematics}, 38(1):207--223.

\bibitem[He et~al., 2011]{He2011Robust}
He, R., Hu, B.~G., Zheng, W.~S., and Kong, X.~W. (2011).
\newblock Robust principal component analysis based on maximum correntropy
  criterion.
\newblock {\em IEEE Transactions on Image Processing}, 20(6):1485--1494.

\bibitem[Kozubowski et~al., 1998]{1998Tails}
Kozubowski, T.~J., Podgorski, K., and Samorodnitsky, G. (1998).
\newblock Tails of levy measure of geometric stable random variables.
\newblock {\em Extremes}, 1(3):367--378.

\bibitem[Liu et~al., 2007]{2007Correntropy}
Liu, W.~F., Pokharel, P.~P., and Principe, J.~C. (2007).
\newblock Correntropy: Properties and applications in non-gaussian signal
  processing.
\newblock {\em IEEE Transactions on Signal Processing}, 55(11):5286--5298.

\bibitem[Miller, 1978]{miller1978properties}
Miller, G. (1978).
\newblock Properties of certain symmetric stable distributions.
\newblock {\em Journal of Multivariate Analysis}, 8(3):346--360.

\bibitem[Petty et~al., 1996]{petty1996freeway}
Petty, K.~F., Noeimi, H., Sanwal, K., Rydzewski, D., Skabardonis, A., Varaiya,
  P., and Al-Deek, H. (1996).
\newblock The freeway service patrol evaluation project: Database support
  programs, and accessibility.
\newblock {\em Transportation Research Part C: Emerging Technologies},
  4(2):71--85.

\bibitem[Santamar{\'\i}a et~al., 2006]{2006Generalized}
Santamar{\'\i}a, I., Pokharel, P.~P., and Principe, J.~C. (2006).
\newblock Generalized correlation function: definition, properties, and
  application to blind equalization.
\newblock {\em IEEE Transactions on Signal Processing}, 54(6):2187--2197.

\bibitem[Wang et~al., 2013]{2013Robust}
Wang, X., Jiang, Y., Huang, M., and Zhang, H. (2013).
\newblock Robust variable selection with exponential squared loss.
\newblock {\em Journal of the American Statistical Association},
  108(502):632--643.

\bibitem[Wang et~al., 2016]{2016Correntropy}
Wang, Y., Tang, Y.~Y., and Li, L. (2016).
\newblock Correntropy matching pursuit with application to robust digit and
  face recognition.
\newblock {\em IEEE Transactions on Cybernetics}, 47(6):1354--1366.

\bibitem[Wu et~al., 2007]{2007Multi}
Wu, Q., Ying, Y., and Zhou, D.~X. (2007).
\newblock Multi-kernel regularized classifiers.
\newblock {\em Journal of Complexity}, 23(1):108--134.

\bibitem[Ying and Zhou, 2007]{2007Learnability}
Ying, Y. and Zhou, D.~X. (2007).
\newblock Learnability of gaussians with flexible variances.
\newblock {\em Journal of Machine Learning Research}, 8:249--276.

\bibitem[Zhou, 2002]{zhou2002covering}
Zhou, D.~X. (2002).
\newblock The covering number in learning theory.
\newblock {\em Journal of Complexity}, 18(3):739--767.

\end{thebibliography}

\end{document}